# DEVELOPING LIGHTWEIGHT DNN MODELS WITH LIMITED DATA FOR REAL-TIME SIGN LANGUAGE RECOGNITION

**SLAIT**

Nikita Nikitin
nikita.nikitin@slait.ai

Eugene Fomin
eugene.fomin@slait.ai

**Abstract**

We present a novel framework for real-time sign language recognition using lightweight DNNs trained on limited data. Our system addresses key challenges in sign language recognition, including data scarcity, high computational costs, and discrepancies in frame rates between training and inference environments. By encoding sign language specific parameters, such as handshape, palm orientation, movement, and location into vectorized inputs, and leveraging MediaPipe for landmark extraction, we achieve highly separable input data representations. Our DNN architecture, optimized for sub **10MB** deployment, enables accurate classification of **343** ASL signs with less than **10ms** latency on edge devices. The data annotation platform 'slait data' facilitates structured labeling and vector extraction. Our model achieved **92%** accuracy in isolated sign recognition and has been integrated into the 'slait ai' web application, where it demonstrates stable inference.

## 1. Introduction

Sign Language Recognition (SLR) and its interpretation into natural language remain significant challenges in machine learning [1]. One approach to addressing this problem is the development of AI models suitable for recognizing human gestures (sign language signs) in a video stream [2]. This task is complicated by the limited availability of training data, the high cost of computational resources [2] and input discrepancies in Frames per Second (FPS) rates between training videos and realtime landmarks stream on the end device. Our previous approach, which relied on Recurrent Neural Network architectures, was limited by the substantial computational and data requirements [3].

In this paper, we briefly outline the results of our research and development efforts in producing lightweight, serverless Deep Neural Networks (DNNs) models trained on limited data that achieve high accuracy and scalability. These models serve as an efficient pre-processing solution for continuous sign language interpretation [4] using Large Language Models. We use one of the most advanced and promising approaches to gesture recognition, similar to sign-to-gloss-to-text (S2G2T) [5], which improves accuracy and reduces the requirements for the volume and quality of training data.

## 2. The Framework Overview

Creating models that recognize hundreds of signs while remaining under 10 MB requires an optimized vector representation of raw video frames. In our approach, we combine modern machine learning techniques with American Sign Language (ASL) academic theory [6]. This approach is based on ASL parameters such as handshape, palm orientation, location, and movement. By leveraging these parameters, we provide well-distinguishable input data suitable for training models on limited datasets, rather than relying on sequences of RGB images or raw landmarks. To extract pose and hand landmark data from each frame, we continue using MediaPipe [7].

The high level of data separability achieved in vector representations based on ASL parameters enables us to develop a lightweight DNN capable of recognizing complex motion gestures and fingerspelling. This DNN can process as few as two input frames while maintaining high accuracy even at a low FPS=5. Additionally, our new augmentation approach generates plausible data, standardizes training videos to the target FPS, and enhances dataset quality through improved labeling and automatic filtering.

Through years of research and development, we have formalized a comprehensive framework that integrates our methodology and technology. This framework encompasses principles for dataset organization and defines the approach to model training. Our low-noise datasets establish a new paradigm for addressing the initial problem. To operate within this paradigm of data scarcity — common in sign languages — we developed our own data annotation platform *"slait.data"*.

## 3. The Data Annotation Platform

The platform *"slait.data"* has been developed for sign language annotation, manual labeling of sample videos, and automated extraction of vector landmarks necessary for dataset creation.

The data preparation pipeline within *"slait.data"* consists of three primary stages. The first stage involves annotation through the description of ASL parameters. The second stage requires the uploading of sample videos for labeling. The third stage enables semi-automated extraction of vector landmarks for continuous storage and export as CSV files for subsequent model training.

The ASL parameter annotation stage entails identifying the features of a unique sign (i.e., human gesture) based on ASL theory that can be described and classified. For each sign, the following features are identified: handshape, handedness and symmetry, lemma (Gloss name), and English translation. An annotator manages the assignment of a unique database ID to each sign to ensure consistency within the dataset by separating different sign variations, merging duplicates, or removing incorrect samples. This process helps minimize the number of classes, thereby reducing information entropy.[1]

Once a sign is described and annotated, video samples acquisition and uploading follows to start labeling. The *"slait.data"* server converts each video file into a sequence of JPG images, and through the user interface, the annotator labels the frames within each video sample. We run MediaPipe [7] on an edge device to extract landmarks.[2] During manual labeling and further quality assurance, we encountered the problem of landmark data loss during MediaPipe processing, which is a significant problem for sign language data acquisition [8].

The labeling process is divided into two stages: data acquisition, and labeling.

Data acquisition. Three data sources have been used:

1. ASL videos produced by SLAIT (60 or 120 FPS; ±50% of the dataset)
2. ASL videos from the ASL Citizen dataset [9] (24 FPS; ±30% of the dataset)
3. ASL videos from various open internet sources (24 or 60 FPS; ±20% of dataset)

Labeling. Annotators use two types of labels for gesture identification. The "Label A" class is assigned to random movements not part of a sign (e.g., hands moving up or down). The "Label B" class marks frames that correspond to actual sign gestures and are intended to be included in the dataset.

We extract 3D landmarks with MediaPipeJS and store them in the vector form. Each vector includes left hand landmarks, right hand landmarks, pose landmarks, video_id, sign_id and fps:

$$< x^l_0, y^l_0, z^l_0, ..., x^l_{20}, y^l_{20}, z^l_{20}, x^r_0, y^r_0, z^r_0, ..., x^r_{20}, y^r_{20}, z^r_{20}, x^p_0, y^p_0, z^p_0, ..., x^p_{24}, y^p_{24}, z^p_{24}, video\_id, sign\_id, fps >$$

As a result, *"slait.data"* builds a database of unique signs and associated CSV files representing the vector landmarks for each class.

### 4. Model Architecture

The architecture of the model is the classifier with branches. We use it to get better vector representation of ASL parameters separately and then make the prediction.

---

[1] To represent collocated signs (e.g., a sequence of unique sign IDs such as CAR + PERSON = DRIVER), the "N-Grammer" algorithm is employed. This algorithm follows the model's output, merging collocated IDs to produce coherent textual output.

[2] Previously, landmark extraction was performed on a server using MediaPipe Python Framework, which compromised the loss of training data. Additionally, discrepancies between the Python and JavaScript implementations of MediaPipe resulted in inconsistencies between training (offline) and production (online) data.

In Figure 1. you can see a simplified schema of the model architecture. Our current biggest model recognizes 343 signs. It was trained with 326,879 samples. It means that after filtering we had about 953 samples per sign.

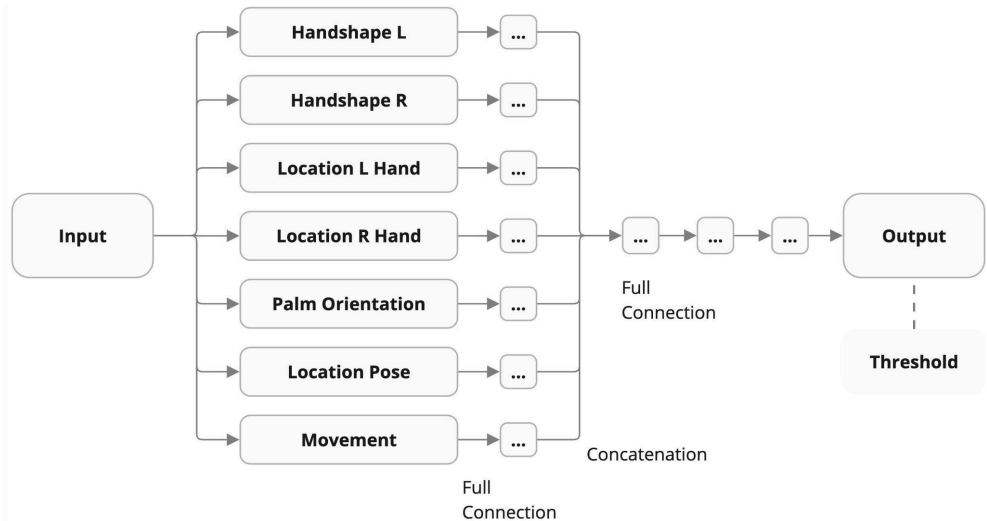

Figure 1.

The input of the model is a vector with the length 947. In includes 7 subvector: "Location L Hand" (63), "Location R Hand" (63), "Location Pose" (75), "Handshape L" (210), "Handshape R" (210), "Palm Orientation" (200), "Movement" (126).

## 5. Model Diagram, Metrics & Hyperparameters
Tensorflow-Keras Plot Model

*(See Supplementary Material, Figure 2.)*

The output of the model is the vector with the length 343. We get it with the help of the Softmax function. For model evaluation we use two accuracy metrics: **Single Frame Sign Recognition (SFSR)** and **Isolated Sign Recognition (ISR).** With SFSR we estimate for what number of frames from the test set the class prediction was correct. With ISR we estimate for what proportion of videos from the test set the class prediction was correct.

**Experimental Setup Snapshot**

| Loss Function | Optimizer | Learning Rate | Weight Decay | Metrics | Epochs | Batch Size |
|---|---|---|---|---|---|---|
| Sparse Categorical Crossentropy | Adam | 0.0001 | 1e-05 | Accuracy | 40 | 64 |

Table 1.

## 6. The Web Application

To reduce SLR error rates while enabling real-time use of our DNN models, we developed *"slait.ai"*, a web-based software that leverages CTC-like algorithms [10]. Its serverless architecture ensures low latency, making it highly effective for applications requiring accurate sign language recognition using only personal computing devices and built-in cameras. A media file is attached [†].

**Summary Table of Results**

| Model Name | Number of ASL signs | Weight | Latency | Accuracy (SFSR) | Accuracy (ISR) |
|---|---|---|---|---|---|
| Basic | 343 | 7.2 mb | 10 ms | 87% | **92%** |

Table 2.

The Basic model (Table 2), trained to recognize 343 unique gestures based on the proposed approach, achieved an ISR accuracy of 92%. The model's weight is approximately 7.2 MB and demonstrated a latency of less than 10 ms in recognizing signs performed in continuous sign language within an edge device video stream, using a MacBook Pro M3 and Snapdragon X Elite.

To prepare the dataset, 12,752 videos were collected, recorded, and labeled using *"slait.data"*, totaling approximately 8 hours of video. These results have been integrated into the web application *"slait.ai"*, which is currently available in beta.

As a result, we designed a novel DNN architecture and a family of models based on it, capable of recognizing isolated sign language with over 2% higher accuracy, using 20 times less data, and supporting 37% more classes than any other models currently available on Kaggle competition [11].

## 7. Conclusion

In future work, we aim to enhance recognition accuracy by supporting various signing styles and expand the vocabulary to 4,000 or more signs, covering the most frequently used ASL lexicon [12]. Additionally, we plan to publish a comprehensive paper that will provide an in-depth description of our framework, detailing its methodologies, optimizations, and real-world applications.

**Acknowledgements**

We are grateful to Adam H. Phillips (Techstars Washington DC, United States) and Ritchie Bryant (Gallaudet University, Washington DC, United States) for their support.

We extend our gratitude to both our former and present colleagues – Antonio Domènech López, Dr. Bill Vicars, Lainie Kleemann, Brenden Gilbert, Felix Santana, and Muhammad Ali – for their contributions.

We acknowledge the efforts of the "slait.data" annotation team, namely – Anna Fomina, Weeklyn B. Curatcho, Ana Lapus, Reyshel R. Rescate, Ayodele Baloye Samuel, and Luke Asher Nuevo–for their accurate work.

Also we are thankful to Jyoti Rai (Research Associate, Institute of Engineering and Technology (IET) of India, Lucknow) for assistance in reviewing this paper.

**References**

[1] Coldewey Devin – *"SLAIT's real-time sign language translation promises more accessible online communication."* – TechCrunch, 2021. URL: https://tcrn.ch/2R0Nd4s

[2] Yulong Li, Bolin Ren, Ke Hu, Changyuan Liu, Zhengyong Jiang, Kang Dang, Jionglong Su KD-MSLRT – *"Lightweight Sign Language Recognition Model Based on Mediapipe and 3D to 1D Knowledge Distillation"*, 2025. URL: https://doi.org/10.48550/arXiv.2501.02321

[3] Domènech López, Antonio – *"Sing Language Recognition - ASL Recognition with MediaPipe and Recurrent Neural Networks"* – UPCommons, October 31, 2020 URL: https://upcommons.upc.edu/handle/2117/343984

[4] Sarah Alyami, Hamzah Luqman – *"A Comparative Study of Continuous Sign Language Recognition Techniques"*, 2024. URL: https://doi.org/10.48550/arXiv.2406.12369

[5] Shahin Nada, Ismail Leila – *"GLoT: A Novel Gated-Logarithmic Transformer for Efficient Sign Language Translation"*, 2025. URL: https://doi.org/10.48550/arXiv.2502.12223

[6] Stokoe, W. C. – *"Sign Language Structure: An Outline of the Visual Communication Systems of the American Deaf"*, 1960. Studies in Linguistics: Occasional Papers 8.

[7] *"Cross-platform, customizable ML solutions for live and streaming media"* – Google, 2025. URL: https://github.com/google-ai-edge/mediapipe?tab=readme-ov-file

[8] Maksym Ivashechkin, Oscar Mendez, Richard Bowden – *"Improving 3D Pose Estimation for Sign Language"* – University of Surrey, August 18, 2023. URL: https://doi.org/10.48550/arXiv.2308.09525

[9] Desai, Aashaka and Berger, Lauren and Minakov, Fyodor O and Milan, Vanessa and Singh, Chinmay and Pumphrey, Kriston and Ladner, Richard E and Daum III, Hal and Lu, Alex X and Caselli, Naomi and Bragg, Danielle – *"ASL Citizen: A Community-Sourced Dataset for Advancing Isolated Sign Language Recognition"*, 2023. URL: https://doi.org/10.48550/arXiv.2304.05934

[10] Sihan Tan, Taro Miyazaki, Nabeela Khan, Kazuhiro Nakadai – *"Improvement in Sign Language Translation Using Text CTC Alignment."* – Institute of Science Tokyo, December 24, 2024. URL: https://doi.org/10.48550/arXiv.2412.09014

[11] *"Google Isolated Sign Language Recognition Competition"* – Kaggle, 2023. URL: https://www.kaggle.com/competitions/asl-signs

[12] *"ASL Signbank"* – Yale, 2024. URL: https://aslsignbank.haskins.yale.edu/

**Media Files:**

[†] Screen_Recording.mp4 URL: https://slait.app/static/Screen_Recording.mp4

# Supplementary Material

# Model Diagram
Tensorflow-Keras Plot Model

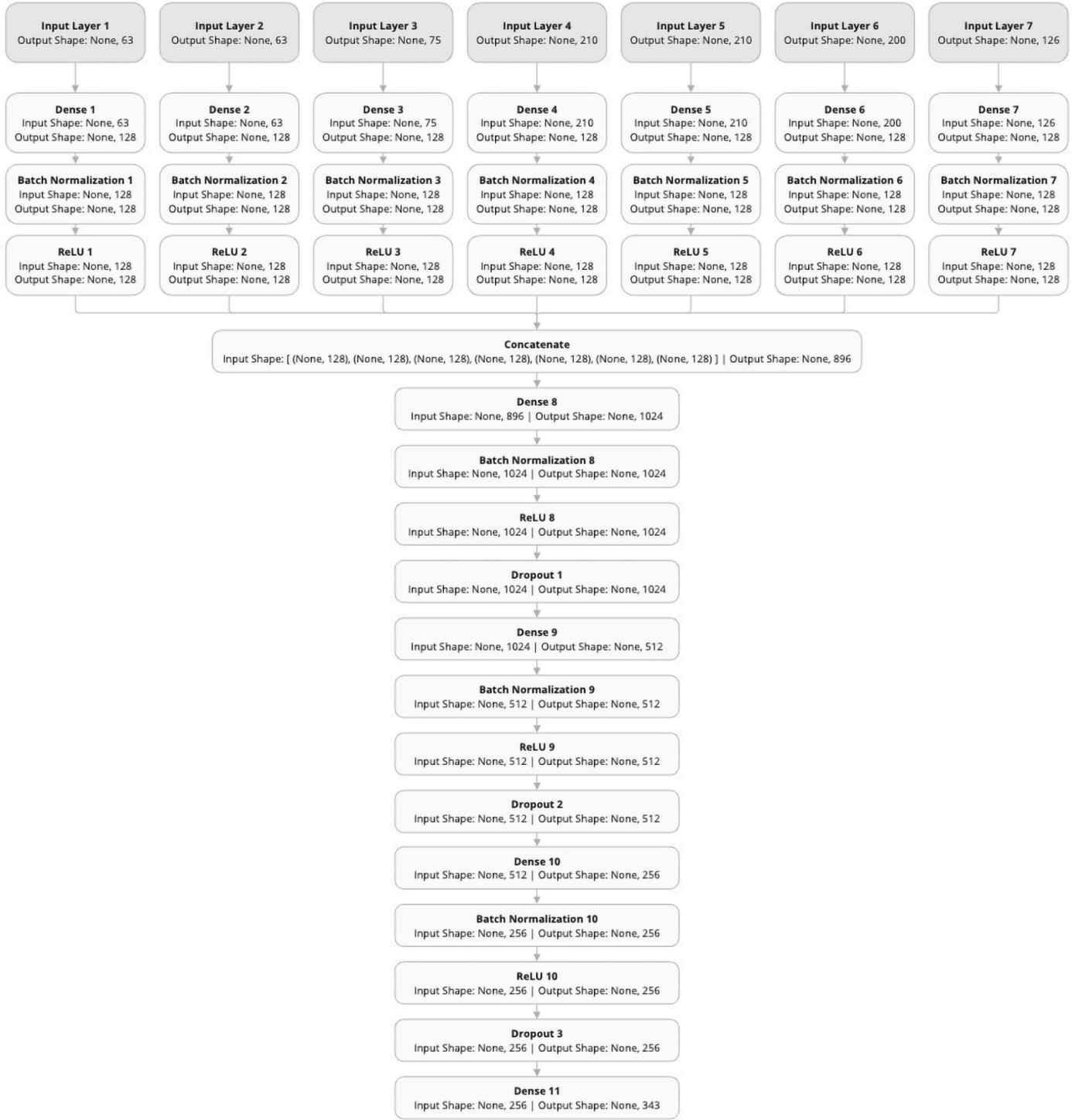

Figure 2.